\documentclass{article}
\usepackage{spconf,amsmath,graphicx,hyperref}
\usepackage{amssymb}
\usepackage{marvosym}
\usepackage[table]{xcolor}
\usepackage{booktabs}
\usepackage{multirow}
\usepackage{tabularx}
\usepackage[most]{tcolorbox}
\usepackage{lipsum} 
\usepackage[normalem]{ulem} 
\usepackage{soul} 
\usepackage[dvipsnames]{xcolor}
\usepackage[font=small,skip=2pt]{caption}

\title{\emph{DSPC}: Dual-Stage Progressive Compression Framework for Efficient Long-Context Reasoning}

\name{Yaxin Gao$^*$\textsuperscript{1,4}\thanks{* Equal contribution} \qquad Yao Lu$^*$\textsuperscript{1,2,4} \qquad Zongfei Zhang\textsuperscript{3} \qquad Jiaqi Nie\textsuperscript{1,4} \qquad Shanqing Yu\textsuperscript{1,4}\textsuperscript{\Letter} \qquad Qi Xuan\textsuperscript{1,4}}
\address{Institute of Cyberspace Security, Zhejiang University of Technology\qquad A*STAR \qquad Amazon\\
Binjiang Institute of Artificial Intelligence, Zhejiang University of Technology}

\begin{document}
%
\maketitle
\begin{abstract}

Large language models (LLMs) have achieved remarkable success in many natural language processing (NLP) tasks. To achieve more accurate output, the prompts used to drive LLMs have become increasingly longer, which incurs higher computational costs. To address this prompt inflation problem, prompt compression has been proposed. However, most existing methods require training a small auxiliary model for compression, incurring a significant amount of additional computation. To avoid this, we propose a two-stage, training-free approach, called \textbf{D}ual-\textbf{S}tage \textbf{P}rogressive \textbf{C}ompression (\emph{DSPC}). In the coarse-grained stage, semantic-related sentence filtering removes sentences with low semantic value based on TF-IDF. In the fine-grained stage, token importance is assessed using attention contribution, cross-model loss difference, and positional importance, enabling the pruning of low-utility tokens while preserving semantics. We validate \emph{DSPC} on LLaMA-3.1-8B-Instruct and GPT-3.5-Turbo under a constrained token budget and observe consistent improvements. For instance, in the FewShot task of the Longbench dataset, \emph{DSPC} achieves a performance of $49.17$ by using only $3\times$ fewer tokens, outperforming the best state-of-the-art baseline LongLLMLingua by $7.76$.

\end{abstract}

\section{Introduction}
\label{sec:intro}
Large language models (LLMs) have achieved remarkable success across many NLP tasks, such as question answering~\cite{wang2025see,izacard2020leveraging}, summarization~\cite{zhang2020pegasus,laban2023summedits}, and code generation~\cite{feng2020codebert,wang2024enhancing}. Carefully designed prompts through in-context learning~\cite{dong2022survey,2022arXiv221210375W}, retrieval-augmented generation~\cite{lewis2020retrieval,li2025multi}, and chain of thoughts~\cite{wei2022chain,zeng2025futuresightdrive} techniques can significantly improve the performance of LLM. However, lengthy prompts come with significant costs, slowing down inference speed, increasing memory cost, and negatively impacting user experience. In addition, overly long prompts often include irrelevant or misleading content that weakens the model’s reasoning.

Consequently, prompt compression has emerged to remove uninformative content while preserving essential semantics. Early methods perform coarse reductions at the phrase or sentence level, which made the compressed text too stiff and lost a lot of useful contextual information~\cite{ghalandari2022efficient}. More recent techniques (e.g., LLMLingua~\cite{jiang2023llmlingua}, LongLLMLingua~\cite{jiang2023longllmlingua} and CPC~\cite{liskavets2025prompt}) enable finer-grained token-level pruning but often require training a small auxiliary compressor model, which adds substantial extra computation.

To address these issues, we propose \emph{DSPC}, a training-free, \textbf{D}ual-\textbf{S}tage \textbf{P}rogressive \textbf{C}ompression framework. Specifically, in the coarse-grained stage, \emph{DSPC} removes sentences with low semantic value based on TF-IDF, reducing the token budget for subsequent processing. In the fine-grained stage, \emph{DSPC} conducts token-level pruning guided by a unified multi-signal token importance score that integrates attention contribution, cross-model loss difference, and positional importance. By considering various token importance scores and adopting a two-stage coarse-to-fine compression mechanism, \emph{DSPC} eliminates the need for additional training and maintains semantic integrity at a high compression ratio with strong generalization.

Our contributions are summarized as follows:
\begin{itemize}
\item 
We propose \emph{DSPC}, a training-free coarse-to-fine prompt compression framework. The first stage removes irrelevant sentences using TF–IDF, and the second stage prunes low-utility tokens based on a unified multi-signal importance score that integrates attention contribution, cross-model loss difference, and positional importance. 

\item Extensive experiments on the LongBench benchmark show that \emph{DSPC} consistently outperforms baseline methods while also achieving faster inference speed, demonstrating the effectiveness and efficiency of our approach.
\item We validate \emph{DSPC} on both open-source models (LLaMA-3.1-8B-Instruct) and closed-source APIs (GPT-3.5-Turbo), as well as across diverse task categories such as question answering, summarization, code generation, and few-shot learning. The consistent improvements confirm that \emph{DSPC} generalizes well across different model scales and application scenarios.
\end{itemize}

\section{PROPOSED METHOD}
\label{sec:format}

\begin{figure*}[htbp]
    \centering
    \includegraphics[width=0.9\textwidth]{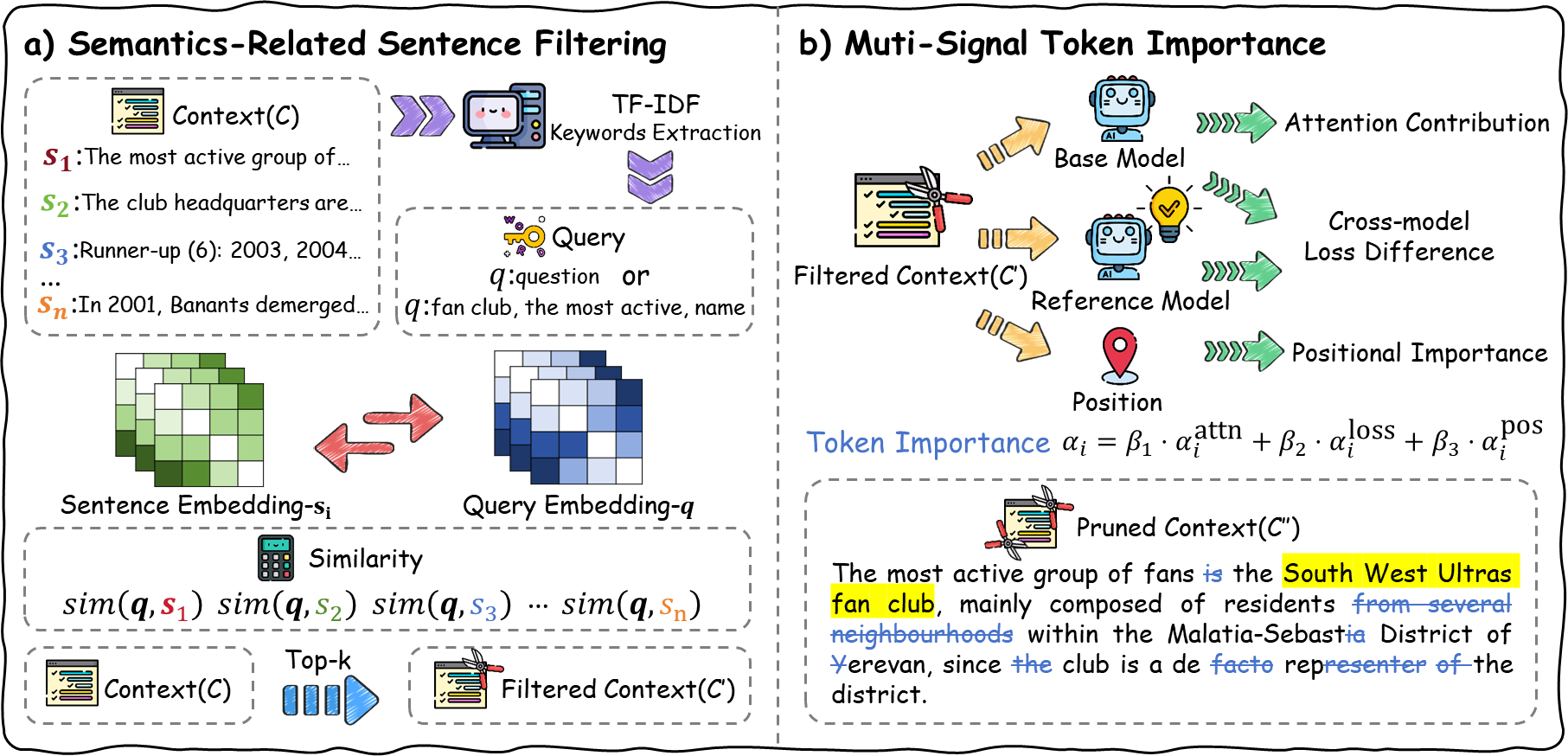}
    \caption{The pipeline of our proposed \emph{DSPC} framework.}
    \label{fig:overview}
\end{figure*}

Here, we elaborate on our two-stage prompt compression method \emph{DSPC}. The overall pipeline is shown in Figure~\ref{fig:overview}. We also provide an example compressed by \emph{DSPC} in Figure~\ref{fig:example}.


\begin{figure}[!t]
    \centering
    \small\begin{tcolorbox}[
        enhanced,
        colback=gray!5!white,      
        colframe=gray!60!black,    
        colbacktitle=gray!60!black,
        coltitle=white,            
        title=Prompt Pruning Process,
        fonttitle=\bfseries,       
        boxrule=1pt,               
        rounded corners=north,     
        arc=1mm,                   
        left=2mm, right=2mm, top=1mm, bottom=1mm
    ]
    Question: What is the name of the most active fan club?\\
    Answer: South West Ultras fan club\\
    \noindent\rule{\linewidth}{0.3pt}
    Stage1: \sout{The current technical director of the academy is the former Russian footballer Ilshat Faizulin.\textbackslash n\textbackslash nFans\textbackslash n\textbackslash n} The most active group of fans is the South West Ultras fan club, mainly composed of residents from several neighbourhoods within the Malatia-Sebastia District of Yerevan, since the club is a de facto representer of the district.\\
    \noindent\rule{\linewidth}{0.3pt}
    Stage2: The most active group of fans \sout{is} the \hl{South West Ultras fan club}, mainly composed of residents \sout{from several neighbourhoods} within the Malatia-Sebast\sout{ia} District of \sout{Y}erevan, since \sout{the} club is a de \sout{facto} rep\sout{resenter} \sout{of} the district.
    \end{tcolorbox}
    \caption{Example of prompt compressed by \emph{DSPC}.}
    \label{fig:example}
\end{figure}

\subsection{Semantics-Related Sentence Filtering}

We begin with a coarse-grained compression step that filters out semantically less relevant sentences from the input context. Formally, the input is segmented into sentences
$
C = \{s_1, s_2, \ldots, s_N\},
$
where each $s_i$ denotes a sentence.
To provide a reference for measuring sentence relevance, we construct a query $q$.  
For tasks with explicit queries (e.g., question answering), $q$ is directly given by the question. 
For tasks without explicit queries (e.g., summarization), we derive $q$ by extracting the top-$k$ keywords from the context $C$ using a TF--IDF-based method~\cite{sparck1972statistical}, defined as:
\begin{equation}
\mathrm{TF\text{-}IDF}(t, C) = \mathrm{TF}(t, C) \cdot \log \left( \frac{N}{\mathrm{DF}(t)} \right),
\end{equation}
where $\mathrm{TF}(t, C)$ is the frequency of term $t$ in $C$, $\mathrm{DF}(t)$ is the number of sentences containing $t$, and $N$ is the total number of sentences.
Both the selected queries and each sentence $s_i$ are encoded into dense vectors using a pre-trained encoder:
\begin{equation}
\mathbf{q}_j = \mathrm{Encoder}(q_j), \quad 
\mathbf{s}_i = \mathrm{Encoder}(s_i).
\end{equation}The relevance score of sentence $s_i$ is then obtained by calculating the cosine similarity between sentence $s_i$ and each query embedding and taking the maximum similarity as the final score:
\begin{equation}
\mathrm{sim}(s_i, q_j) = \frac{\mathbf{q}_j \cdot \mathbf{s}_i}{\|\mathbf{q}_j\| \,\|\mathbf{s}_i\|}.
\end{equation}
\begin{equation}
\mathrm{score}(s_i) = \max\; \{\mathrm{sim}(s_i, q_j)\}, \quad j=1,\dots,k.
\end{equation}
Subsequently, sentences are ranked by their scores in descending order, and the top-$k$ sentences are selected as the filtered context:
\begin{equation}
C' = \mathrm{Top}\text{-}k \big(\mathrm{score}(s_i) \big),
\end{equation}
where $C' \subseteq C$, $k = \lfloor \rho N \rfloor$, and $\rho \in (0,1]$ controls the proportion of sentences retained.

\subsection{Multi-Signal Token Importance}

The sentence-filtered context $C'$ may still contain redundant or less informative tokens. 
To further compress the input, we perform a fine-grained, token-level pruning step. 
Specifically, we compute the final token importance score based on three complementary signals: attention contribution, cross-model loss difference, and positional importance. This token-level pruning retains semantically important tokens while removing less relevant ones, thereby preserving key information and continuity in the compressed input.

Let $C' = \{s_1, s_2, \ldots, s_{N'}\}$ be the sentence-filtered context from the previous stage, 
where $N' = \lfloor \rho N \rfloor$. Each sentence $s_i$ is then split into tokens using the tokenizer, obtaining the token sequence
$
T = \{t_1, t_2, \ldots, t_n\}.
$ 
To determine which tokens to keep, we assign an importance score to each token $t_i$, which is calculated from three factors: attention contribution, cross-model loss difference, and positional importance. We describe these in detail below.

\textbf{Attention Contribution.} 
We use the self-attention weights from the last layer of the base model to measure the token importance, as they naturally reflect the model's assessment of which tokens are most relevant in the context.
Let $\mathbf{A} \in \mathbb{R}^{H \times n \times n}$ be the attention weight, where $H$ is the number of attention heads. The attention contribution $\alpha_i^{\text{attn}}$ is computed as the average attention received by token $t_i$ from all other tokens:
\begin{equation}
\alpha_i^{\text{attn}} = \frac{1}{H\times n} \sum_{h=1}^{H} \sum_{j=1}^{n} A_{h,j,i}.
\end{equation}
Attention contribution highlights which tokens the model focuses on.

\textbf{Cross-model Loss Difference.} Attention contribution identifies which tokens the model focuses on, but it alone cannot capture inter-model learning disparities that reveal each token’s actual contribution to model learning, which is crucial for assessing token importance~\cite{lin2024rho}. To this end, we further introduce a cross-model token-wise loss difference that captures information not reflected in attention patterns. For each token $t_i$, let $p^{m}(t_i \mid t_{<i})$ denote the probability assigned by model $m \in \{\text{base}, \text{ref}\}$ given its preceding context $t_{<i}$. 
The token-wise loss is defined as the negative log-likelihood:
\begin{equation}
\ell_i^{m} = - \log p^{m}(t_i \mid t_{<i}).
\end{equation}
The token importance is then computed as the difference between the base and stronger reference models:
\begin{equation}
\alpha_i^{\text{loss}} = \ell_i^{\text{base}} - \ell_i^{\text{ref}}.
\end{equation}
A large positive value of $\alpha_i^{\text{loss}}$ corresponds to the reference model predicting $t_i$ with much higher confidence than the base model, indicating that the token carries more information content for the task and needs to be retained.


\textbf{Positional Importance.} 
As observed in~\cite{liu-etal:2023:arxivlostinthemiddle}, transformers typically focus more on the beginning and end of long sequences, while ignoring the content in the middle. To mitigate this bias and better represent tokens in the middle of a sequence, we incorporate positional information to correct for it. Specifically, we use a fixed Gaussian function to assign a positional importance to each token based on its sequence index:
\begin{equation}
\alpha_i^{\text{pos}} = 1 + \lambda \cdot \exp\Big(-\frac{(i - n/2)^2}{2\sigma^2}\Big),
\end{equation}
where $n/2$ denotes the sequence center, $\lambda$ and $\sigma$ are fixed constants that determine the magnitude and spread of the positional boost. This Gaussian weighting increases the importance of tokens near the center while maintaining a baseline contribution for boundary tokens, effectively mitigating potential mid-sequence information loss.

\textbf{Token Selection.}
After obtaining $3$ importance signals (attention contribution, cross-model loss difference and positional importance) for each token, we combine them into a final score $\alpha_i$:
\begin{equation}
\alpha_i = \beta_1 \cdot \alpha_i^{\text{attn}} + \beta_2 \cdot \alpha_i^{\text{loss}} + \beta_3 \cdot \alpha_i^{\text{pos}},
\end{equation}
where $\beta_1$, $\beta_2$, and $\beta_3$ control the relative contributions of each factor. 
All tokens in $C'$ are then ranked according to their final importance scores and the top-$k$ tokens are selected to form the compressed token-level context:
\begin{equation}
C'' = \mathrm{Top}\text{-}k \big( \{t_i \in C'\}, \ \alpha_i \big),
\end{equation}
where $k = \lfloor \delta \, n \rfloor$ and $\delta \in (0,1]$ is the desired compression ratio. Finally, we get $C''$, which preserves the semantics of the original paragraph without affecting the performance.

\begin{table*}[t] 
\centering
\small
\caption{Performance comparison of different prompt compression methods at different compression ratios on the LongBench dataset. Since LLMLingua uses the officially recommended GPT-2 model, which does not support Chinese, its tokens and $1/\tau$ are represented by "--". $1/\tau$ represents the overall compression ratio, that is, how many times the original input is reduced relative to the compressed content.}
\setlength{\tabcolsep}{10pt}
\renewcommand{\arraystretch}{1}
\begin{tabular}{l|cccccc|ccc}

\toprule
\multirow{2}{*}{\textbf{Methods}}
& \multicolumn{9}{c}{\textbf{LongBench}} 
\\
\cline{2-10}
& SingleDoc & MultiDoc & Summ. & FewShot & Synth. & Code & AVG & Tokens & $1/\tau$ \\

\midrule
\multicolumn{10}{c}{\textbf{2,000 tokens constraint}} \\

\midrule
\rowcolor{gray!10} 
LLMLingua    & 7.45 & 8.42 & 17.70 & 9.77 & 4.53 & 21.52 & 11.57 &- &- \\
LLMLingua-2   & 16.14 &\textbf{16.24} & 18.11 & 12.66 & 22.32 & 25.39 & 18.48 &1,838 &5x \\
\rowcolor{gray!10} 
LongLLMLingua  & 12.43 & 12.57 & 18.14 & 12.57 & 8.12 &27.85 & 15.28 & 1,855 &5x \\

\rowcolor{gray!25} \textbf{\emph{DSPC}} 
                     & \textbf{17.84} & 15.94 & \textbf{22.29} & \textbf{29.04} 
                     & \textbf{36.10} &\textbf{37.69} & \textbf{26.48}  & 1,597 &5x \\ 

\midrule
\multicolumn{10}{c}{\textbf{3,000 tokens constraint}} \\
\midrule
\rowcolor{gray!10} 
LLMLingua         & 12.47 & 7.92 & 17.79 & 8.44 & 4.12 & 20.86 & 11.93 &- &- \\
LLMLingua-2       & \textbf{20.63} & 14.68 & 18.43 & 11.45 & 29.81 & 24.84 & 19.97 &2,571  &3x \\
\rowcolor{gray!10} 
LongLLMLingua     & 18.30 & 13.86 & 18.51 & 10.94 &16.03 & 26.10 & 17.29 &2,719  &3x \\

\rowcolor{gray!25} \textbf{\emph{DSPC}} 
                    & 19.30 &\textbf{16.48}  &\textbf{22.68}  & \textbf{30.52}
                    &\textbf{34.43}  &\textbf{38.57}  &\textbf{26.99} & 2,318 &3x\\
\bottomrule
\end{tabular}
\label{tab:longbench_compression}
\end{table*}

\begin{table}[t]
\centering
\caption{Performance comparison of \emph{DSPC} and other baselines on FewShot task of LongBench under a 3,000 token limit. We use GPT-3.5-Turbo for evaluation.}
\setlength{\tabcolsep}{2pt} 
\renewcommand{\arraystretch}{0.95}
\resizebox{\columnwidth}{!}{%
\begin{tabular}{c|c|*{4}{>{\centering\arraybackslash}p{1.8cm}}}
\toprule
\multirow{2}{*}{\textbf{Dataset}} & \multirow{2}{*}{\textbf{Tasks}} 
& \multicolumn{4}{c}{\textbf{GPT-3.5-Turbo (3,000 tokens)}} \\
\cline{3-6}
 & &\footnotesize LLMLingua &\footnotesize LLMLingua2 &\footnotesize LongLLMLingua &\footnotesize \emph{DSPC} \\
\midrule
\multirow{2}{*}{\textbf{Longbench}}
&\cellcolor{gray!10}FewShot
& \cellcolor{gray!10}44.71 & \cellcolor{gray!10}42.17 & \cellcolor{gray!10}41.41 & \cellcolor{gray!10}\textbf{49.17} \\
&Synth.
& 6.18 & 39.85 & 17.20 & \textbf{46.56} \\

\bottomrule
\end{tabular}%
}
\label{tab:fewShot_gpt}
\end{table}

\begin{table}[t]
\centering
\footnotesize  
\caption{Impact of different hyperparameters on the performance of LLaMA-3.1-8B-Instruct on SingleDoc.}
\setlength{\tabcolsep}{4pt} 
\renewcommand{\arraystretch}{0.95} 
\begin{tabular}{c|c|c|c|c|c} 
\toprule
\multirow{2}{*}{\textbf{Token Budget}} &\textbf{Stage 1} & \multicolumn{3}{c|}{\textbf{Stage 2}} & \multirow[c]{2}{*}{\textbf{Performance}} \\
\cline{2-5}
&\scriptsize Compression Ratio &\scriptsize $\beta_1$ &\scriptsize $\beta_2$ &\scriptsize $\beta_3$ &  \\ 
\cmidrule(lr){1-6}
\multirow{8}{*}{\textbf{3,000}} &\cellcolor{gray!10}0.8 &\cellcolor{gray!10}0.6 &\cellcolor{gray!10}0.3 &\cellcolor{gray!10}0.1 & \cellcolor{gray!10}17.32 \\
&\textbf{0.7} & \textbf{0.6} & \textbf{0.3} & \textbf{0.1} & \textbf{18.23} \\
&\cellcolor{gray!10}0.6 & \cellcolor{gray!10}0.6 & \cellcolor{gray!10}0.3 & \cellcolor{gray!10}0.1 & \cellcolor{gray!10}18.06 \\
&0.7 & 0.1 & 0.1 & 0.8 &  17.39\\
&\cellcolor{gray!10}0.7 & \cellcolor{gray!10}0.8 & \cellcolor{gray!10}0.1 & \cellcolor{gray!10}0.1 &  \cellcolor{gray!10}16.79\\
&0.7 & 0.1 & 0.8 & 0.1 &  17.39\\

&\cellcolor{gray!10}0.7 & \cellcolor{gray!10}0.5 & \cellcolor{gray!10}0.3 & \cellcolor{gray!10}0.2 &  \cellcolor{gray!10}18.21\\
&0.7 & 0.4 & 0.5 & 0.1 &  18.11\\
\bottomrule
\end{tabular}
\label{tab:ablation_small}
\end{table}

\begin{table}[t]
\centering
\footnotesize  
\caption{Inference speed of LLaMa-3.1-8B-Instruct on the Fewshot task of the LongBench dataset.}
\setlength{\tabcolsep}{4pt} 
\renewcommand{\arraystretch}{0.95} 
\begin{tabular}{c|c|cc} 
\toprule
\textbf{Dataset} &\textbf{Method} &\textbf{Inference Time} &\textbf{Speedup} \\
\midrule
\multirow{4}{*}{\textbf{FewShot}} &\cellcolor{gray!10}LLMLingua &\cellcolor{gray!10}4.69s &\cellcolor{gray!10}1.3x \\
                        &LLMLingua2 & 5.15s & 1.2x \\
                        &\cellcolor{gray!10}LongLLMLingua &\cellcolor{gray!10}4.83s &\cellcolor{gray!10}1.3x\\
                        &\emph{DSPC} & 4.59s & 1.3x\\
\bottomrule
\end{tabular}
\label{tab:speedup}
\end{table}

\section{EXPERIMENT}
\label{sec:pagestyle}
\subsection{Experimental Setup}
\textbf{Models.} 
We use \textbf{LLaMA-3.2-1B-Instruct} as the base model to calculate the attention contribution, 
\textbf{LLaMA-3.2-3B-Instruct} as the reference model for computing cross-model loss difference.
We further use larger models (e.g., \textbf{LLaMA-3.1-8B-Instruct} and \textbf{GPT-3.5-Turbo}) for evaluation.


\textbf{Datasets.} 
We evaluate \emph{DSPC} on \textbf{LongBench}~\cite{bai2023longbench}, a benchmark of long-context tasks. 
It covers single- and multi-document question answering (SingleDoc, MultiDoc), summarization (Summ.), few-shot learning (FewShot), synthetic reasoning (Synth.), and code tasks (Code), providing a comprehensive evaluation of long-context capabilities.

\textbf{Baselines.} 
We compare our method against several state-of-the-art prompt compression baselines, including \textbf{LLMLingua}, \textbf{LLMLingua2} and \textbf{LongLLMLingua}. To ensure a fair comparison, we utilize the same models and datasets. All models share the same tokenization and pre-processing.

\textbf{Hyperparameters.} 
In sentence filtering, $\rho \in (0,1]$ controls the proportion of semantic proportion.
In token-level pruning, hyperparameters $(\beta_1, \beta_2, \beta_3)$ balance the weight coefficients of attention contribution, cross-model loss difference, and positional importance.

\subsection{Results}
We evaluate the effectiveness of \emph{DSPC} on LongBench using LLaMA-3.1-8B-Instruct. As shown in Table~\ref{tab:longbench_compression}, \emph{DSPC} consistently outperforms baselines with input lengths limited to $2k$ and $3k$ tokens. Under the $3k$ token limit,  \emph{DSPC} achieves improvements of $+1.33$, $+1.80$, $+4.17$, $+19.07$, $+4.61$ and $+12.48$ on SingleDoc, MultiDoc, Summ., FewShot, Synth., and Code, respectively, over the second-best baseline. Furthermore, under the stricter $2k$ token limit, gains are still impressive compared to the second-best baseline, with $+1.71$ on SingleDoc, $+4.15$ on Summ., $+16.39$ on FewShot, $+13.76$ on Synth., and $+9.82$ on Code. To further evaluate the generalization ability of \emph{DSPC}, we use GPT-3.5-Turbo for evaluation under the limit of $3k$ token. As shown in Table~\ref{tab:fewShot_gpt}, \emph{DSPC} still outperform all baselines. These experiments demonstrate the effectiveness and generalizability of \emph{DSPC}.


\textbf{Ablation Experiments.} We explore the impact of different hyperparameter settings on the performance of LLaMA-3.1-8B-Instruct on the SingleDoc task. As shown in Table~\ref{tab:ablation_small}, different hyperparameter settings have a certain impact on model performance. When the compression ratio of the first stage is $0.7$ and $\beta_1=0.6$, $\beta_2=0.3$, $\beta_3=0.1$, the model achieves the best performance. In contrast, a higher first-stage compression ratio leads to performance degradation, indicating that balancing information retention across stages is crucial to improving model performance.

 
In addition to ablation experiments, we also measured inference speedup. Specifically, we test the inference time of LLaMa-3.1-8B-Instruct on the Fewshot task of the LongBench dataset and calculate the average inference time for a single sample. As shown in Table~\ref{tab:speedup}, our method achieves the best inference speedup, demonstrating that \emph{DSPC} improves the inference speed while maintaining performance.

\section{Conclusion}
\label{sec:majhead}
In this paper, we introduce \emph{DSPC}, a training-free, dual-stage prompt compression framework that first applies TF-IDF for sentence filtering and then leverages attention contribution, cross-model loss difference and positional importance for fine-grained token pruning. Extensive experiments on LongBench using LLaMA-3.1-8B-Instruct and GPT-3.5-Turbo demonstrate consistent performance improvements and faster inference speed compared to existing baselines.



\clearpage
\bibliographystyle{IEEEbib}
\bibliography{strings,refs}

\end{document}